\documentclass{article}
\usepackage{interspeech2014, url, graphicx, amsmath, cite, balance}
\usepackage[font={small}]{caption}

\usepackage{tikz}
\usetikzlibrary{arrows,shapes,backgrounds,positioning,fit}
\tikzstyle{ctext} = [rectangle,inner sep = 4pt]

\newif\ifniko
\nikotrue

\def\fig#1{Fig.~\ref{fig:#1}}
\def\sec#1{Section~\ref{sec:#1}}
\def\tab#1{Table~\ref{tab:#1}}
\def\eq#1{(\ref{eq:#1})}

\let\underscore=\_
\def\_{\checkmath_\underscaore}
\def\checkmath#1#2{\ifmmode\def\next##1{#1{\rm##1}}\else\let\next=#2\fi\next}

\def\math#1{\relax\ifmmode#1\else$#1$\fi}

\def\a{\hat a}

\def\mincdet{\math{C\_{det}^{\rm min}}}

\def\cllr{\math{C\_{llr}}}

\def\C{{\cal C}}
\def\S{{\cal S}}
\def\P{{\cal P}}
\def\H{{\cal H}}
\def\l{{\ell}}
\def\r{{r}}
\def\Lnorm{L\_{norm}}
\def\Hcross{H\_{cross}}
\def\barHcross{{{\overline H}\_{cross}}}
\def\barHcrossmin{{{\overline H}\_{cross}^{\rm min}}}

\ninept

\def\fudge{0.5}
\bigskipamount=\fudge\bigskipamount
\medskipamount=\fudge\medskipamount
\smallskipamount=\fudge\smallskipamount
\makeatletter
\renewcommand{\section}{\@startsection
  {section}
  {1}
  {}
  {-\bigskipamount}%
  {0.5\bigskipamount}%
  {}}%

\renewcommand{\subsection}{\@startsection
  {subsection}%
  {2}%
  {}%
  {-\medskipamount}%
  {0.5\medskipamount}%
  {}}%

\renewcommand{\subsubsection}{\@startsection
  {subsubsection}%
  {3}%
  {}%
  {-\smallskipamount}%
  {0.5\smallskipamount}%
  {}}%
\makeatother

\renewenvironment{thebibliography}[1]{%
  \begin{oldthebibliography}{#1}%
    \setlength{\parskip}{0.3ex}%
    \setlength{\itemsep}{0.3ex}%
  }%
  {%
  \end{oldthebibliography}%
}

\title{Constrained speaker linking}
\name{David A. van Leeuwen$^1$ and Niko Br\"ummer$^2$}
\address{$^1$Netherlands Forensic Institute, The Hague and Radboud University Nijmegen, The Netherlands\\
$^2$AGNITIO Research, Somerset West, South Africa}

\begin{document}

\maketitle
\begin{abstract}
In this paper we study speaker linking (a.k.a.\ partitioning) given constraints of the distribution of speaker identities over speech recordings.  Specifically, we show that the intractable partitioning problem becomes tractable when the constraints pre-partition the data in smaller cliques with non-overlapping speakers.  The surprisingly common case where speakers in telephone conversations are known, but the assignment of channels to identities is unspecified, is treated in a Bayesian way.  We show that for the Dutch CGN database, where this channel assignment task is at hand, a lightweight speaker recognition system can quite effectively solve the channel assignment problem, with 93\,\% of the cliques solved.  We further show that the posterior distribution over channel assignment configurations is well calibrated.  

\end{abstract}

\section{Introduction}
\label{sec:introduction}

Speaker linking~\cite{speaker-linking:2010}, also known as speaker partitioning~\cite{niko-odyssey:2010}, is the general problem of finding which utterances in a collection of recordings are spoken by the same speaker.  It  resembles speaker diarization~\cite{Tranter:2006}, in that usually no prior training material of any of the speakers is given, but in speaker diarization there is the additional problem of segmentation, with the possibility of overlapping speech~\cite{Huijbregts:2009a}.  In~\cite{niko-odyssey:2010} it is shown that the task covers many different problems in the area of speaker recognition, ranging from traditional `NIST SRE-style' speaker detection to speaker counting and unsupervised adaptation.  The authors approach the problem in a purely Bayesian way, meaning that the task is defined as computing a posterior distribution over all possible speaker partitions, given the data and a prior distribution over the partitions.  Because of the combinatoric explosion of the number of partitions, the Bayesian approach can only be taken for small-sized problems.  A different approach to essentially the same problem was taken in \cite{speaker-linking:2010}, where a solution was sought in terms of agglomerative clustering, i.e., making sequences of speaker linking decisions, concentrating on large scale problems.  Speaker linking has found application in large scale speaker diarization~\cite{Huijbregts:2011a, Ferras:2012, Campbell:2012}, where the task of speaker diarization within a single recording is extended to finding the same speaker across multiple recordings, and to very long recordings for which the speaker diarization problem becomes computationally challenging~\cite{Huijbregts:2011a}.  

In this paper, we are investigating how prior information in the speaker linking problem can be used.  We will use a probabilistic approach following~\cite{niko-odyssey:2010}, and analyse prior information in terms of uncertainty.  Then, we will apply this to a specific, but remarkably common situation in telephone speech data collections.  This is the case where speaker identities in telephone conversations are known, but the assignment of identities to the two channels in the recorded conversation is unknown.  We refer to this problem as the \emph{channel assignment} task.  In The Netherlands, we have seen this situation more than once.  In the telephone interception recordings made in police investigations, both parties of a conversation are recorded in separate channels.  The `identity' of the speaker in such a case is limited to technical metadata such as the telephone number or the IMSI and IMEI numbers of the SIM and handset, respectively.  This metadata is stored in the interception database, but for a reason unknown to us, it is impossible to tell which identity is recorded in which channel of the 2-channel audio file.  From a forensic point of view, this does not pose a problem.  A specific person is under investigation, and this means that calls to/from this person's telephone are intercepted.  When in the content of the call there is incriminating evidence, it is that fragment that is going to be important.  If the speaker identity of that fragment is questioned, forensic speaker comparison is going to play a role in the case.  But even then, the channel assignment of the phone number is not important.  This is different when we want to employ data mining approaches to all recordings in a police investigation.  Then the link between identity and channel is very relevant indeed.  

A second example is the data collection `Dutch Spoken Corpus' (\emph{Corpus Gesproken Nederlands}, CGN~\cite{Oostdijk:2003}).  This is a large general speech database of contemporary Dutch spoken in the Netherlands and Belgium, with a wide variety of speech sources and speaking styles, and annotated at various levels of details.  As such, it is widely used by researchers in linguistics, language and speech technology in these countries.  In 2008, the  data was used as training material in the project N-Best, an evaluation of Dutch speech recognition systems~\cite{N-Best:2009}.  At the preparation stage it turned out that, despite the multitude of annotations available with CGN, the mapping of orthographic annotation (and speaker identities) to channels within a telephone conversation was unknown.  A possible reason for this is that the telephone conversations recorded as two channels only contribute to a small portion of the entire CGN, and that despite the very broad application scenario included in the design of CGN, requirements for automatic speaker recognition may not have been fully worked out.  At the time, additional manual speaker attribution was made and distributed to the participants of N-Best.  Participants reported different strategies for dealing with the originally missing speaker-to-channel annotation~\cite{limsi-nbest:2009, Huijbregts:2009}.  

Other situations where speaker identities are known, but the exact mapping is unknown are large scale diarization problems.  E.g., in~\cite{Ferras:2012}, a sequence of meetings is processed with diarization, and later the speaker linking between meetings is carried out.  Sometimes the speaker information is partially known, as in the case of the speaker diarization of broadcast shows where TV-guide metadata can reveal the names of some of the speakers in the shows~\cite{diarization-retrieval:2011}.   

\section{Reducing uncertainty}
\label{sec:reducing-uncertainty}

We are going to describe the speaker linking problem in terms of uncertainty, where both prior information and speaker recognition can contribute to lowering the uncertainty in the speaker identities in the database.  We will do this in terms of the total entropy in the database, parameterized by size and other prior information.  The number of recordings will be denoted by $2M$, anticipating that we are going to investigate a collection of $M$ conversations, each contributing two separate recordings, shortly.  In the following, we will consecutively add constraints, or prior information, to the analysis. 

\subsection{Speaker-homogeneous recordings}
\label{sec:speak-hom}

The overall restriction in this paper is that each speech recording contains speech from only a single speaker.  This excludes the tasks of speaker diarization.  The speaker entropy is $H\_I = \log B_{2M}$, where $B_n$ is the $n$th `Bell number'~\cite{niko-odyssey:2010}.  Here we have applied no further priors to the number of speakers and their distribution, not even that---for a telephone call---a speaker can't talk to herself. 

\subsection{Number of speakers is known}
\label{sec:num-speak}

When additionally the number of speakers~$N$ in the database is known, the entropy is reduced to $H\_{II} = \log {2M\atopwithdelims\{\}N}$, where $n\atopwithdelims\{\}k$ denotes the Stirling number of the second kind.  An upper bound approximation is $2M\log N - \log N!$, i.e., for each recording the identity can be any of $N$ speakers, and we must compensate for the arbitrary speaker labelling. 

\subsection{Telephone conversations}
\label{sec:tel-con}

When we know that the conversations are all telephone conversations, we can exclude situations where one speaker occurs in both sides of the conversation.  This reduces the entropy just a little further by approximately $M\log \frac N{N-1}$.

\subsection{Speakers in conversations are known}
\label{sec:speakers-known}

We now make a big step to the channel assignment task described in the introduction: the speaker identities of the parties participating in a conversation are known, but it is unknown in which channel which speakers is.  This prior results in the reduction of entropy to~$H\_{IV} = M\log 2$.  If the entropy is expressed in bits, we have $H\_{IV} = M$. 

\subsection{Application of speaker recognition}

Speaker linking can further reduce the entropy.  For instance, if a single speaker occurs in all $M$ conversations, and there are $M+1$ speakers in total, it is easy to see that, if speaker recognition works flawlessly, the entropy can be reduced to~$0$. However, there are many different partitionings of the speakers over the calls possible, and not all will have te same potential w.r.t.\ entropy reduction even with a perfect speaker recognition system.  For instance, if the same two speakers occur in all $M$ conversations, there is still $\log 2$ entropy left.  And in the extreme situation that there are $2M$ speakers in the database, the entropy remains at $M\log 2$ as before.   

The entropies described in the previous paragraph are \emph{potentially attainable} entropies in the case of perfect speaker linking.  However, speaker recognition is based on statistical models of speech and speakers, and can make errors.  We therefore need a different measure to compute the effect of speaker recognition, and we will use the cross entropy for that. 

\section{Speaker recognition}

We will use de detection capability of a speaker recognition system as the information used in speaker linking.  We assume that when two utterances (different conversation sides) $x$ and $y$ are available, the recognizer can provide a log-likelihood-ratio for comparing the two
\begin{equation}
  \label{eq:llr}
  \lambda(x,y) = \log \frac{P(x,y \mid \H_1)}{P(x,y\mid \H_2)}
\end{equation}
where $\H_1$ and $\H_2$ are the hypotheses that $x$ and $y$ are spoken by the same or different speakers, respectively.  We assume that the system is \emph{well calibrated}, i.e., that $\lambda$ can be used effectively to compute a minimum risk Bayes decision. 

We will utilize the speaker detector for analyzing various sub-partitionings of the telephone channel assignment problem.  We will have to introduce some notation at this point.  A \emph{conversation labelling} is the way speaker pairs are distributed over telephone conversations, i.e., without explicitly encoding which speaker is in which channel.  If such a conversation labelling encompasses $M$ conversations, there are $2^M$ possible configurations~$\C$ of the speaker pairs over the channels in the conversations.  These can be numbered in binary notation, using $\l\ $ and $\r$ instead of traditional 0's and 1's.  The channel assignment problem of \sec{speakers-known} is to find the correct configuration given the conversation labelling and the speech data.  

For a given configuration~$\C$ of speakers over recordings, the likelihood is $L_\C = P(X\mid\C)$, where $X$ denotes the relevant speech.  Given the prior probability for this configuration $\pi_\C$ and the likelihoods, the posterior can be computed using
\begin{equation}
  \label{eq:posterior}
  P(\C \mid X) = \frac{\pi_\C L_\C}{\sum_{\C'} \pi_{\C'} L_{\C'}},
\end{equation}
where the summation is over all $2^M$ possible configurations~$\C'$.  For the stated task, it is not unreasonable to set the prior uniformly at $\pi_i = 2^{-M}$.

In \tab{entropies} we have summarized the way the reduction of entropy evolves by specifying more constraints.  As a numerical example, we have used the test data in the experiment explained in \sec{experiments}. 

\begin{table}
  \centering
  \caption{Some example values for the entropies of the speaker paritioning problem. We have used the parameters of the test data from CGN (348 conversations involving 356 speakers) as an example of the reduction in entropy~$H$, which is expressed in bits here. $F$ is the average confusion, see~\sec{evaluation-metrics}.}
  \label{tab:entropies}
  \medskip
  \begin{tabular}{|l|c|c|c|c|}
    \hline
    Sec-& Expression& \multicolumn2{|c|}{CGN}\\
    tion& $H$& $H$& $F$\\
    \hline
    \ref{sec:speak-hom}& $H\_{I} = \log B_{2M}$ & 4163.2& 3991\\
    \ref{sec:num-speak}& $H\_{II} = \log {2M\atopwithdelims\{\}N}$ & 3292.7 & 704\\
    \ref{sec:tel-con}&$H\_{III} = H\_{II} - M\log\frac N{N-1}$& 3291.3& 702\\
    \ref{sec:speakers-known}& $H\_{IV} = M\log 2$& 348& 1.0\\
    \hline
  \end{tabular}
  \vskip-\baselineskip
\end{table}

\subsection{Single speaker chains}
\label{sec:single-speak-chains}

\begin{figure}
  \begin{tikzpicture}[thick]
    \def\a{$\l\vphantom{\r}$}
    \def\b{$\r\vphantom{\l}$}
    \node[ctext](row1){$a$ \& $b$};
    \node[ctext,below=10pt of row1](row2){$a$ \& $c$};
    \node[ctext,right=10pt of row1](A11){\a};
    \node[ctext,right=5pt of A11](B11){\b};
    \node[ctext,below=10pt of A11](A21){\a};
    \node[ctext,below=10pt of B11]{\b};
    \draw[-] (A11) to (A21);
    \node[ctext,right=40pt of A11](A12){\a};
    \node[ctext,right=40pt of B11](B12){\b};
    \node[ctext,below=10pt of A12](A22){\a};
    \node[ctext,below=10pt of B12](B22){\b};
    \draw[-] (A12) to (B22);
    \node[ctext,right=90pt of A11](A13){\a};
    \node[ctext,right=90pt of B11](B13){\b};
    \node[ctext,below=10pt of A13](A23){\a};
    \node[ctext,below=10pt of B13]{\b};
    \draw[-] (B13) to (A23);
    \node[ctext,right=140pt of A11](A14){\a};
    \node[ctext,right=140pt of B11](B14){\b};
    \node[ctext,below=10pt of A14](A24){\a};
    \node[ctext,below=10pt of B14](B24){\b};
    \draw[-] (B14) to (B24);
  \end{tikzpicture}
  
  \caption{The four linking configurations for a single speaker chain in 2 conversations.  The link indicates the channel of $a$.}
  \label{fig:diagram}
  \vskip-\baselineskip
\end{figure}
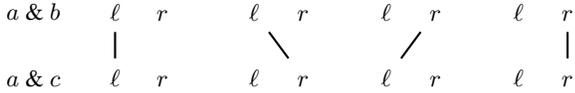

Let us first consider the simplest case, a set of two conversations with a single common speaker.  We denote this `target speaker' by~$a$, occurring twice, with conversation partners $b$ and $c$, respectively.  There are four possible configurations over the two channels $\l,\r$ for the conversations~1 and~2, namely $(ab,ac)$, $(ab,ca)$, $(ba,ac)$ and $(ba,ca)$, cf.~\fig{diagram}.  Using $\Lnorm$ to denote the likelihood that all recordings have different identities, the log-likelihood for the first in the four configurations is 
\begin{equation}
  \label{eq:2}
  \log L_1 = \log \Lnorm + \lambda(1\l,2\l), 
\end{equation}
with similar likelihoods $L_2, \ldots, L_4$ for the other configurations.  Note that with \eq{posterior}, the factor $\Lnorm$ cancels in the posteriors~$P_i$.  A maximum posterior linking decision would therefore correspond to a clustering step based on maximum likelihood.  

Next, we will consider a situation with $M$ conversations, each with the same `target speaker'~$a$, and all different conversation partners.  Using $\Lnorm^{2M}$ to denote the likelihood that all $2M$ recordings are from different speakers, a first approximation to the likelihood~$\tilde L_1$ for the first configuration $(ab, ac, \ldots, aM)$ could be computed using
\begin{equation}
  \label{eq:5}
  \log \tilde L_1 = \log \Lnorm^{2M} + \sum_{i=1}^{M-1}\lambda(i\l,(i+1)\l),
\end{equation}
i.e., making the ``link'' between assumed speakers $a$ in channel~$\l$ from 1 to~2, from 2 to~3, etc.  But this is just one possible linking, and there are in fact $M\choose 2$ possible log-likelihood-ratios, from which a chain of $M-1$ need to be chosen.  A better approximation therefore is to include them all, and scale these to $M-1$ contributions:
\begin{equation}
  \label{eq:multiple-link}
  \log L_1 = \log \Lnorm^{2M} + \frac2M \sum_{i<j} \lambda(i\l, j\l).
\end{equation}

\subsection{Re-occurring conversation partners}

Thus far we have considered that case that all conversation partners of a particular target~$a$ are different.  However, it is likely that in some conversations the same partner~$b$ occurs.  This leads to two different cases:
\begin{description}
\item[unresolvable] Target~$a$ and $b$ are always paired in the database.  In this case, a relative linking is possible, but the absolute attribution of the speakers to the links cannot be resolved.
\item[resolvable] There is more than one conversation partner for $a$ (or $b$).  In principle, the speaker linking could be resolved by speaker recognition. 
\end{description}
If we further concentrate on the subset of the database involving speaker~$a$, then multiple occurrences of $b$ can add to the log-likelihood.  The additional contribution of speaker $b$ to the log-likelihood of the 1st configuration is, similar to \eq{multiple-link}
\begin{equation}
  \label{eq:other-link}
  \frac2{M_b}\sum_{i<j} \lambda(i\r, j\r)
\end{equation}
where the sum is over all $M_b$ conversations involving $b$ as a conversation partner of $a$---for the first configuration $b$ is in channel~$\r$.  This average log-likelihood ratio should be added to $L_1$ of \eq{multiple-link}, and similar terms to the log-likelihoods of the other configurations. 

\subsection{Independent cliques}
\label{sec:independent-cliques}

If there are two sets of speakers~$\S_1$ and $\S_2$ in the database that do not share a single conversation, we call these different \emph{cliques}.  When $\C_1$ is a configuration within all conversations~$X_1$ involving $\S_1$, and $\C_2$ correspondingly for $\S_2$, then the posterior
\begin{equation}
  \label{eq:3}
  P(\C_1,\C_2\mid X_1, X_2) = P(\C_1\mid X_1) P(\C_2 \mid X_2),
\end{equation}
i.e., the configurations can be treated independently, reducing the total number of configurations that need to be computed.  If the design of the database does not allow for the analysis of separate cliques, the problem of computing the posterior distribution over all configurations of the database will still be intractable for large~$M$. 

\section{Experiments}
\label{sec:experiments}

In this section we apply the constrained speaker linking approach sketched above to `NL/component c' of the Dutch CGN database~\cite{Oostdijk:2003}, which consists of telephone conversations between acquaintances in The Netherlands, recorded using a telephony platform.  As mentioned before this database actually lacks the channel-to-speaker assignment, but within the context of the N-Best project~\cite{N-Best:2009} this information had been manually added.  We treat this part of CGN as a prototypical example of the channel assignment task, and use the manually added reference for evaluation purposes.  It consists of 352 conversations (704 sides) involving 357 different speakers.  

The speaker recognition system we use is a lightweight `UBM/GMM dot-scoring' system~\cite{sdv-sre:2008} implemented entirely in the new high performance language for numerical computing Julia.\footnote{\url{http://julialang.org/}}  We use a standard acoustical front end based on 20 MFCC's plus first and second derivatives, energy-based Speech Activity Detection, and 4 second feature warping~\cite{Pelecanos:2001}.  The 1024 component UBM was trained gender-independently on `NL/component d' (telephone conversations recorded using a mini-disc, 600 conversations, 176 speakers, 21 hours).  The speaker comparison score is a linear approximation to the MAP-adapted~\cite{Gauvain:1994a,Reynolds:2000} GMM / UBM log likelihood ratio score.  For probabilistically interpretable likelihood ratios~\eq{llr} we self-calibrated the collection of test scores using CMLG~\cite{score-distr:2013} before applying \eq{posterior}.\footnote{This procedure is comparable to determining \mincdet\ in speaker detection evaluation.}  The equal error rate of the system evaluated on a full score  matrix of all recordings of the test data is $E_= = 7.0\,\%$. 

NL/component~c of CGN consist of many independent cliques, presumably as a result of way the speakers were recruited.  There are 125 cliques of 2--5 speakers, each having 1--6 conversations amongst each other.  During the course of this study, the recognition system hinted on potential labelling errors.  We manually listened to the conversations of cliques to check the labelling.  We started with the largest cliques of 6 conversations, selecting only those with the biggest log error, and worked our way down in clique size.  In most cases labeling errors were very clear, observable from gender, mentioned names and relations, as many cliques seemed to have been formed within families.  We corrected the labeling of suspicious cliques once, without further feedback from the speaker recognition and linking system.  As a result of various dubious speaker labels, we ended up using 348 conversations involving 356 speakers. 

\subsection{Evaluation metrics}
\label{sec:evaluation-metrics}

A natural metric that extends the earlier mentioned entropy considerations is the cross entropy.  The cross entropy between the evaluator's and recognizer's posteriors for a clique $\S$ with a true partitioning $\P$ is 
\begin{align}
  \Hcross &= -\sum_{\C\in\S} P(\C \mid \P) \log P(\C \mid X)\\
  &= -\log P(\P \mid X),
\end{align}
i.e., $\Hcross$ can be seen as a logarithmic scoring rule, which is known to be strictly proper~\cite{DeGroot:1983}.  Because $\Hcross$ covers $M_\S$ conversations, the average cross entropy per conversation is 
\begin{equation}
  \label{eq:6}
  \barHcross = \frac\Hcross{M_\S}. 
\end{equation}
An intuitive meaning of this average entropy is through the perplexity, $\exp \barHcross$, as the average number of configurations to choose from for a set of conversations.  Along the lines of the Albayzin Language Recognition Evaluation metric~\cite{Rodriguez:2013}, we will rather use the \emph{confusion}
\begin{equation}
  \label{eq:8}
  F = e^\barHcross - 1. 
\end{equation}
It represents the average number of \emph{wrong} alternatives.  $F$ is 0 for a system operating perfectly.  In \tab{entropies} we have tabulated the reduction of $F$ based on the prior entropy.  In the last row, which represents our channel assignment task, there is just one configuration alternative to the correct one.  The goal is to further reduce this to 0 by applying speaker linking.  

Finally, we define the clique error rate~$E$ as the average number of cliques for which the maximum posterior configuration is not the true speaker configuration. 

\subsection{Linking}
\label{sec:linking}

In \tab{results} the results of the speaker attribution experiment for CGN are shown.  We have conditioned the performance on the complexity~$|\C|=2^{M_\S}$, the number of configurations in the speaker attribution task, thereby averaging over cliques with the same number of configurations.  For a clique of one conversation, $|\C|=2$, speaker linking cannot resolve the uncertainty, $P_{1,2} = 0.5$, and the entropy is 1 bit.  There are 8 such speaker pairs that both occur only once in the database, and a further four cases with $|\C|=4$ where the same speakers are paired twice.  We will further discard these unresolvable cases in the analysis. 

\begin{table}
  \centering
  \caption{Performance of the linking experiment on CGN, Dutch, component c, grouped by complexity.  $|\C|$ is the number of configurations for a clique, $N_\C$ is the number of cliques of this complexity.  Entropies are measured in bits.}
  \label{tab:results}
  \smallskip
  \begin{tabular}{|c|r|l|l|c|}
    \hline
    $|\C| = 2^{M_\S}$& $N\_\C$& $\barHcross$& $F$& $E$ (\%)\\
    \hline
    2&   8& 1.0&  1.0& NA\\
    4&  61& 0.205& 0.153& 3.3\\
    8&  13& 0.201& 0.150& 7.7\\
    16& 26& 0.032& 0.023& 7.7\\
    32&  9& 0.041& 0.029& 22\\
    64&  5& 0.019& 0.013& 20\\
    \hline
    all resolvable& 110& 0.078& 0.056& 7.0\\
    \hline
  \end{tabular}
  \vskip-\baselineskip
\end{table}

The average cross entropy drops below 1 bit for all complexities above 2, which is an increasing reduction w.r.t.\ the original entropy $\bar H = 1$.  The confusion drops steadily with increasing complexity, and finally, partitioning error rate is more or less stable at about 7\,\% on average.  We have computed $N_\C$-weighted averages over the resolvable cliques in the last row of the table. 

\subsection{Calibration}

The scores $\lambda$ have been `self-calibrated' using a linear transform $\lambda = as+b$, where $s$ is the score and $a$ and $b$ are parameters found to minimize the 2-class cross entropy (e.g., \cllr) in a classical detection set-up, quite similar to logistic regression.  For the posterior~\eq{posterior} the offset~$b$ cancels, so effectively $a$ is the only parameter of importance in this task. 

We want to investigate if the approximation by averaging of log likelihoods as in \eq{multiple-link} and the addition of independent evidence as in \eq{other-link} is not making the system over-confident.  Therefore we re-calibrate the scaling factor $a$ for different clique complexities, and investigate the change in $a$ and the improvement in average cross entropy per conversation. 

\begin{table}
  \centering
  \caption{The effect of calibration.  $\protect\barHcross$ is the average cross entropy before re-calibration, cf~\tab{results}.  The data are taken only over resolvable cliques.}
  \label{tab:calibration}
  \smallskip
  \begin{tabular}{|l|c|c|c|c|}
    \hline
    $M_\S$& $\barHcross$& $\barHcrossmin$& $a\_{min}/a$\\
    \hline
    2& 0.082& 0.071& 1.48\\
    3& 0.201& 0.177& 0.65\\
    4& 0.032& 0.029& 1.42\\
    5& 0.041& 0.040& 0.77\\
    6& 0.019& 0.018& 0.79\\
    \hline
    all& 0.078& 0.076& 1.17\\
    \hline
  \end{tabular}
  \vskip-\baselineskip
\end{table}

In \tab{calibration} the improved $\barHcrossmin$ for the different complexities is shown, with the additional factor $a\_{min}/a$ applied to the recognizer's log likelihoods.  These values of $a\_{min}/a$ vary a bit, but given the relatively small numbers of cliques that are involved in the optimization, this may be expected.  The optimization over all resolvable cliques results in a value $a\_{min}/a=1.17$, which is very close to unity.  This means that the original calibration based on detection trials only was good, and that the averaging operation~\eq{multiple-link} and addition of evidence from other links in the clique~\eq{other-link} do not make the posteriors overconfident. 

\section{Conclusions}
\label{sec:conclusions}

We have seen in the channel assignment task that with higher complexity, i.e., higher prior entropy, the average cross entropy per conversation quickly drops towards zero.  On the one hand, with increasing complexity the task gets harder because there are more configurations to discriminate between.  On the other hand, with higher complexity the increased clique size provides more opportunity for the recognition system to determine the correct configuration with confidence.  These two effects more-or-less cancel in the fraction of correctly found configurations.  With a very lightweight speaker recognizer, trained with a small amount of domain specific speech material, the channel assignment task can be carried out with a high degree of success.  This is indeed a complete Bayesian solution to the speaker partitioning problem as proposed in~\cite{niko-odyssey:2010}, which is deemed intractable in general.  In cases like the CGN database, where conversation clique sizes~$M_\S$ are relatively small, the computational load still is negligible (computing the posteriors for all cliques, given all relevant log-likelihood-ratios, takes about 1 second), but of course the load grows exponentially with $M_\S$.  However, it is likely that also in forensic investigations the cliques are small so that we expect that our approach can be used there as well. 

\clearpage

\vsize=0.54\vsize

\bibliographystyle{IEEEtran}
\bibliography{david-bibdesk}

\balance

\end{document}